\documentclass[letterpaper, 10 pt, conference]{ieeeconf}  

\IEEEoverridecommandlockouts                              

\usepackage{amsmath}
\usepackage{bm}
\usepackage{amsfonts}
\usepackage{multirow}
\usepackage{graphicx}
\usepackage{subfig}
\usepackage{tikz}
\usepackage{cite}
\usepackage{subfig}
\usepackage{algorithm2e}
\usepackage[font=footnotesize]{caption}
\usepackage[noend]{algpseudocode}
\usepackage{hyperref,cleveref}
\usepackage[colorinlistoftodos,prependcaption,textsize=tiny]{todonotes}

\newlength\Fcolumnseprule
\newcommand{\Ereject}{E_\mathrm{reject}}
\newcommand{\jointPC}{\mathcal{P}_j}
\newcommand{\point}{\mathbf{p}}

\title{\LARGE \bf
CorAl -- Are the point clouds Correctly Aligned?
}

\author{Daniel Adolfsson, Martin Magnusson, Qianfang Liao, Achim J. Lilienthal, Henrik Andreasson 
  \thanks{The authors are with the MRO lab of the AASS research centre
    at \"Orebro University, Sweden.
    E-mail: \texttt{Daniel.Adolfsson@oru.se}}
  \thanks{This work has received funding from the Swedish Knowledge Foundation (KKS) project ``Semantic Robots'' and European Union's
    Horizon~2020 research and innovation programme under grant
    agreement No 732737 (ILIAD)     and 101017274 (DARKO).
    \newline 978-1-6654-1213-1/21/\$31.00 \textcopyright 2021 IEEE}%
}

\begin{document}

\maketitle
\thispagestyle{empty}
\pagestyle{empty}

\begin{abstract}

In robotics perception, numerous tasks rely on point cloud registration. However, currently there is no method that can automatically detect misaligned point clouds reliably and without environment-specific parameters. We propose ``CorAl'', an alignment quality measure and alignment classifier for point cloud pairs, which facilitates the ability to introspectively assess the performance of registration. CorAl compares the joint and the separate entropy of the two point clouds. The separate entropy provides a measure of the entropy that can be expected to be inherent to the environment. The joint entropy  should therefore not be substantially higher if the point clouds are properly aligned.
Computing the expected entropy  makes the method sensitive also to small alignment errors, which are particularly hard to detect, and applicable in a range of different environments. 
We found that CorAl is able to detect small alignment errors in previously unseen environments with an accuracy of 95\% and achieve a substantial improvement to previous methods.


\end{abstract}
\section{Introduction}

In order to create safe and efficient mobile robots, introspective and reliability-aware capabilities are required to assess and recover from perception failures. Many perception tasks, including localization~\cite{adolfsson_submap_2019}, scene understanding and sensor calibration~\cite{della_corte_unified_2019}, rely on point cloud registration. However, registration may provide incorrect estimates due to local minima of the registration cost function~\cite{9013051}, uncompensated motion distortion~\cite{zhang_loam_2014}, noise or when the registration problem is geometrically under-constrained~\cite{8462890,softconstraints}. 
Consequently, it is essential to measure alignment quality and to reject or re-estimate alignment when quality is low.
In the past, an extensive number of methods have been proposed to assess the alignment quality of point cloud pairs \cite{icp,Segal_2009,gimlop,HavrdaCharvatTsallis,fuzzybnb,aoki2019pointnetlk,evangelidis2018joint,bouaziz2013sparse,Stoyanov2012ijrr,rusinkiewicz-2001-fasticp,7335490}. These metrics can typically be used to measure a relative alignment error in the process of registration, but provide limited information on whether the point clouds are correctly aligned once registration has been carried out~\cite{Bogoslavskyi2017AnalyzingTQ}. 
Until today, few studies have targeted the measurement of alignment correctness after registration~\cite{Almqvist,Bogoslavskyi2017AnalyzingTQ} and previous works report that alignment correctness classification based on AdaBoost and NDT score function decrease when applied to point clouds acquired from new environments~\cite{Almqvist}.
In this paper, we propose ``CorAl'' (Correctly Aligned?): 
A method to introspectively measure and detect misalignment between previously registered point cloud pairs.
CorAl specifically aims to bridge the gap in classification performance when applied to new unseen environments.



\begin{figure}
    \centering
    {\includegraphics[trim={0.3cm 0cm 0.0cm 0cm},clip,width=\hsize]{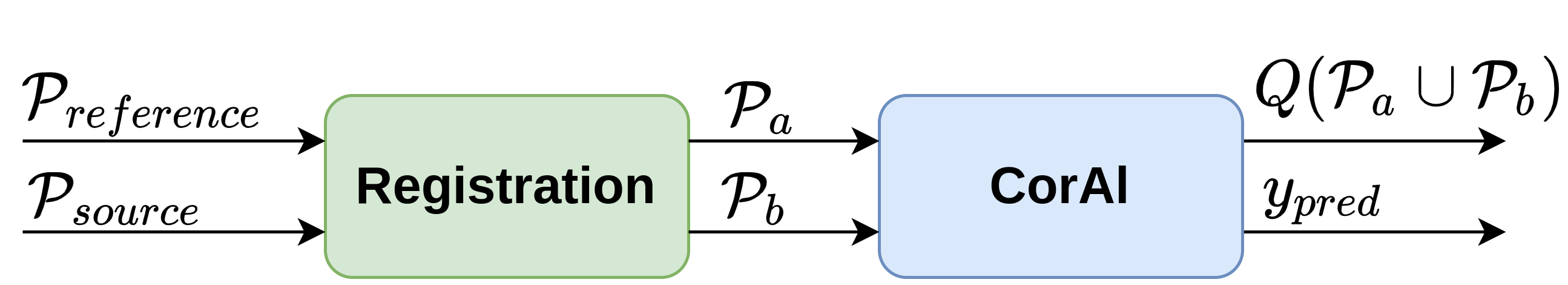}\label{fig:coral_overview}}
    \caption{CorAl depicted in blue, operates on a pair of 
    point clouds $\mathcal{P}_a,\mathcal{P}_b$ and can classify misalignment ($y_{pred}$) by comparing the differential entropy in the point clouds separately and jointly. Additionally, CorAl outputs a per-point quality measure $Q(\mathcal{P}_a\cup\mathcal{P}_b)$ that highlights misaligned parts.}
    \label{fig:my_label}
\end{figure}

\begin{figure}
\vspace{-0.5cm}
\centering
\subfloat[][\raggedright  $\mathcal{P}_a$ colored by entropy.]{\includegraphics[trim={20.0cm 5cm 20.0cm 5cm},clip,width=0.49\hsize]{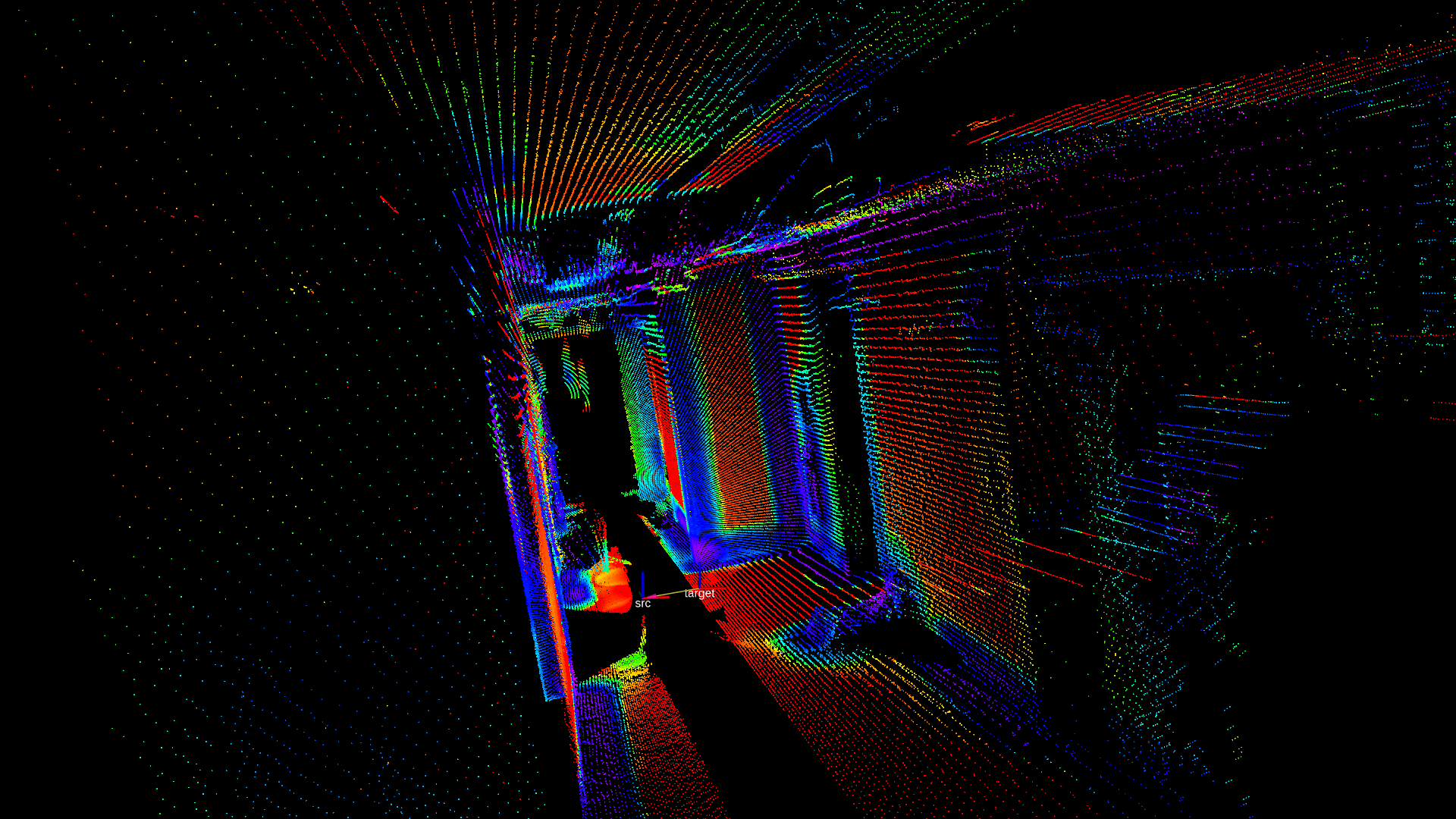}\label{fig:entropy_src}}
\subfloat[][\raggedright $\mathcal{P}_b$ colored by entropy.]{\includegraphics[trim={20.0cm 5cm 20.0cm 5cm},clip,width=0.49\hsize]{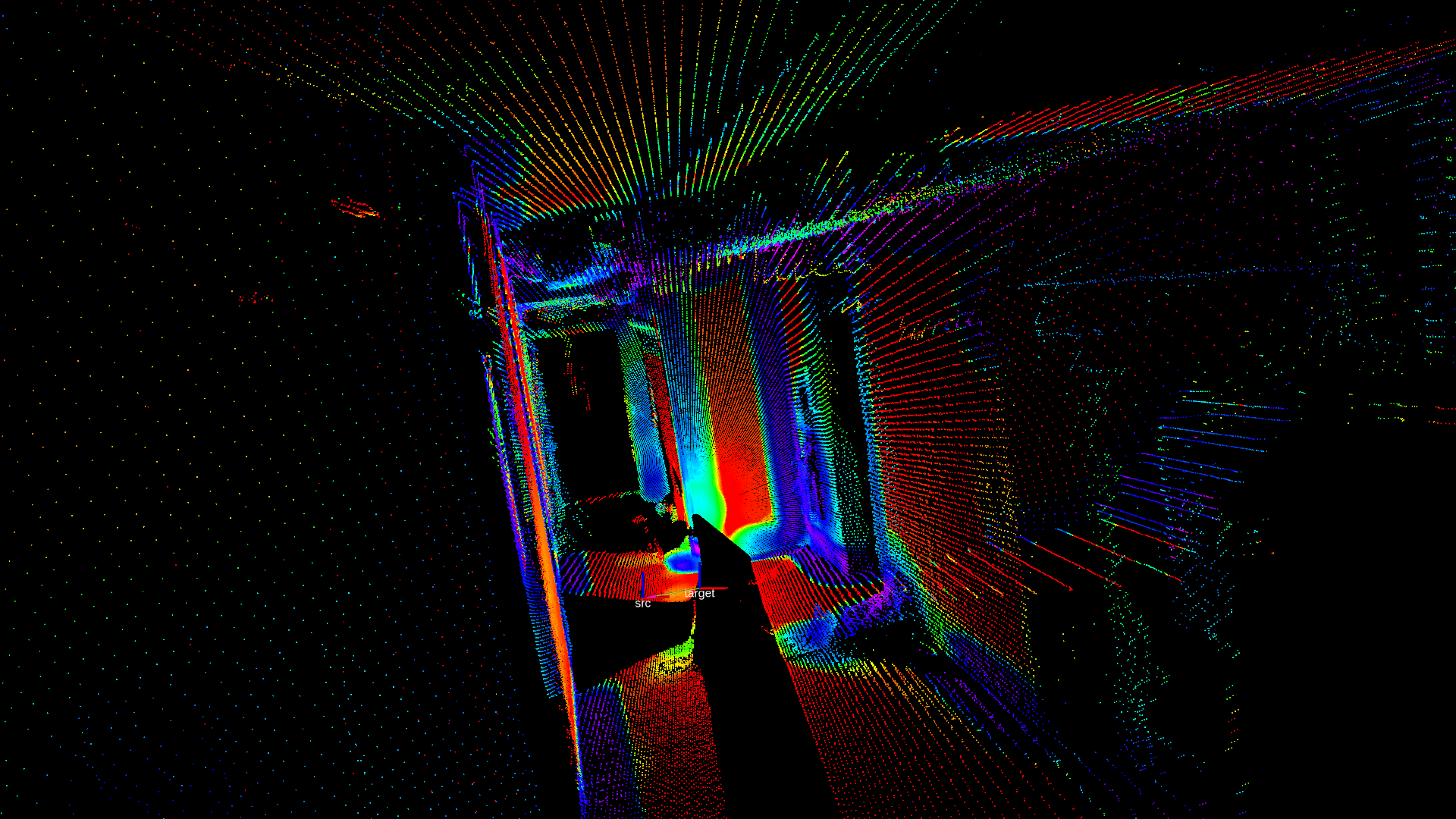}\label{fig:entropy_target}}
\\
\vspace{-0.2cm}
\subfloat[][\raggedright Correctly aligned $\mathcal{P}_a \cup \mathcal{P}_b$ colored by quality measure.]{\includegraphics[trim={20.0cm 5cm 20.0cm 5cm},clip,width=0.49\hsize]{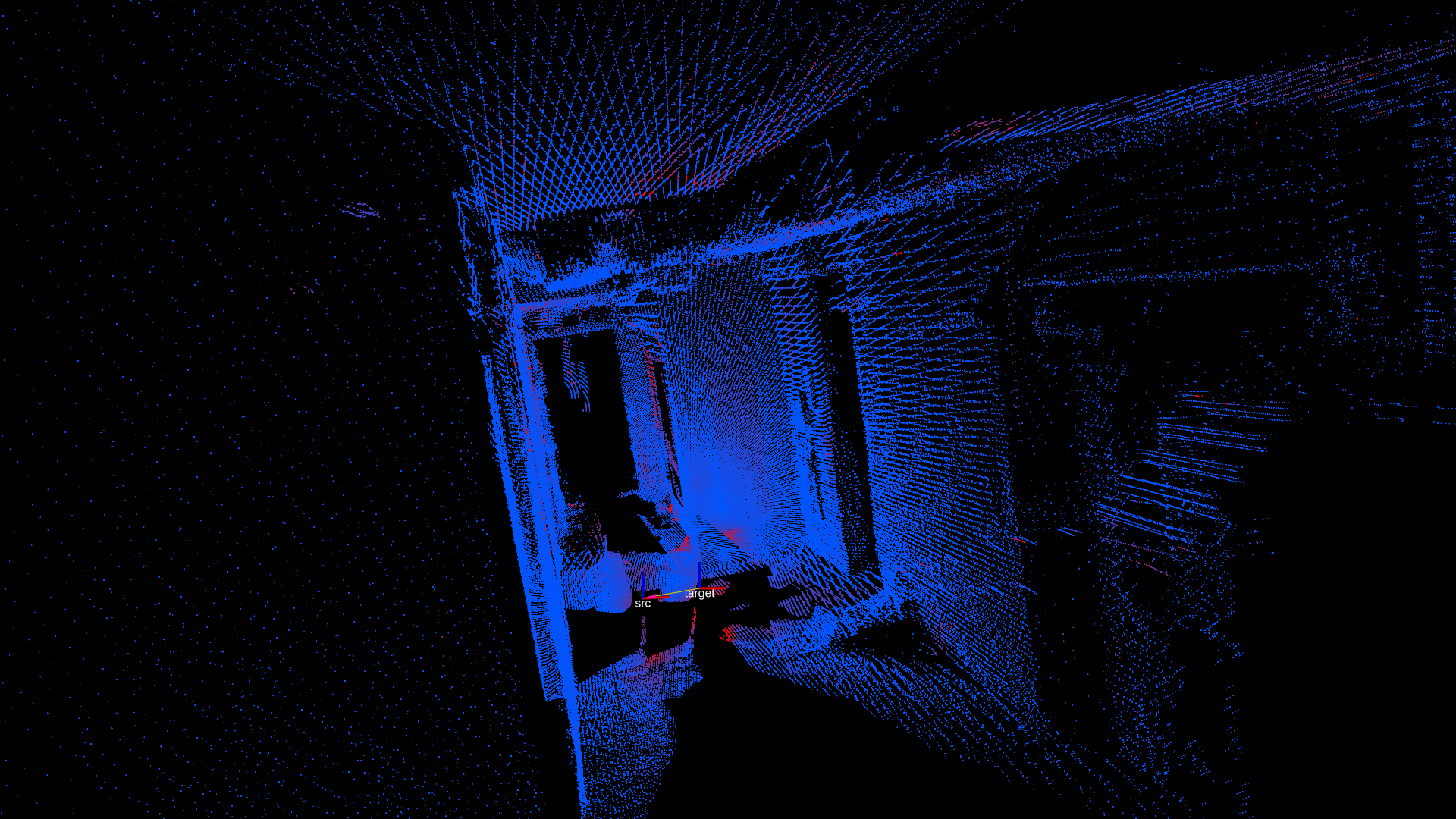}\label{fig:entropy_diff_aligned}}
\subfloat[][ Misaligned $\mathcal{P}_a \cup \mathcal{P}_b$ colored by quality measure.]{\includegraphics[trim={20.0cm 5cm 20.0cm 5cm},clip,width=0.49\hsize]{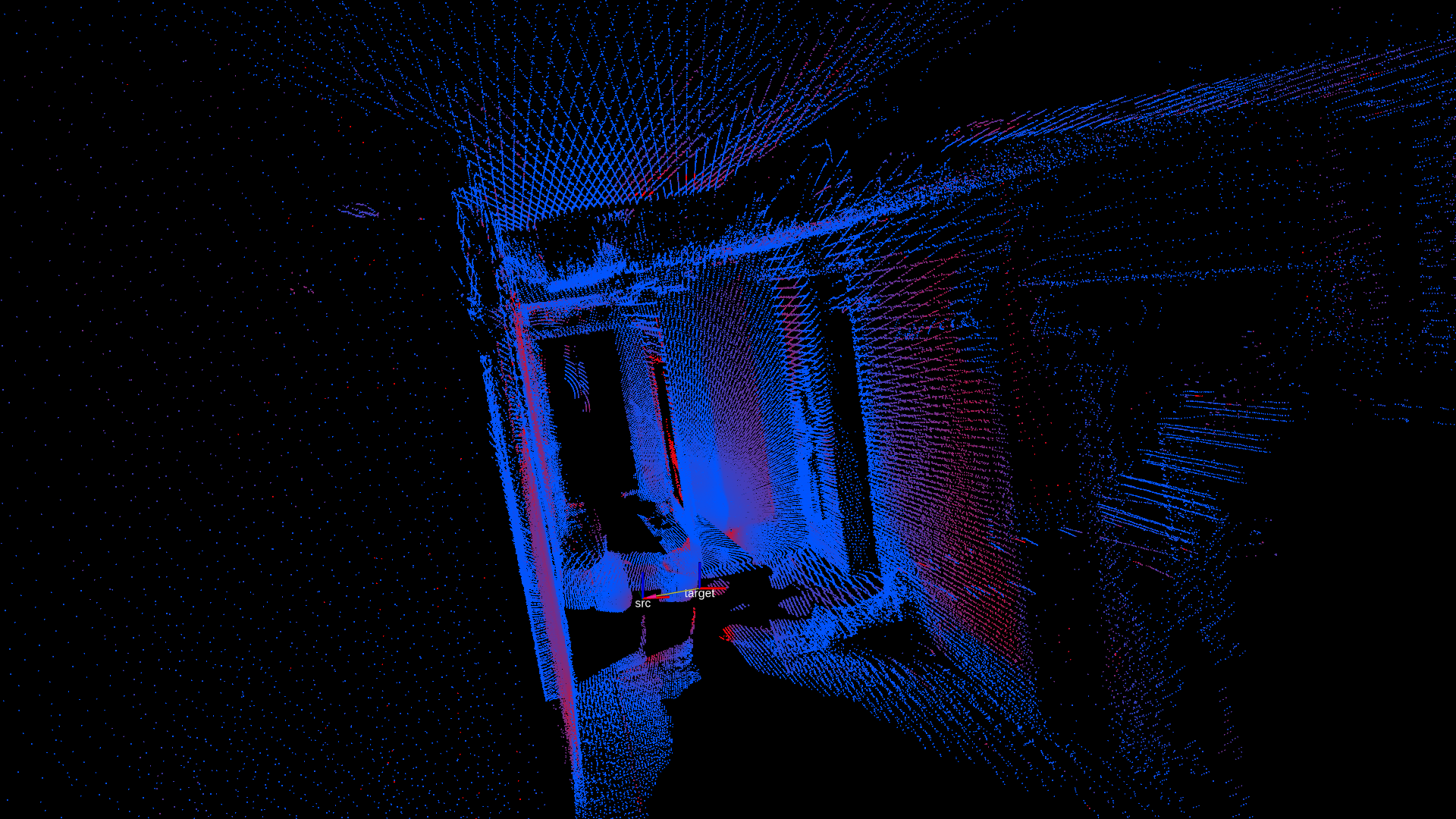}\label{fig:entropy_diff_misaligned}}
\caption{\label{fig:diff_entropy} Top: Differential entropy in point clouds separately. Bottom: The joint point cloud ($\mathcal{P}_a \cup \mathcal{P}_b$) colored by per-point quality measure $q_k(\mathcal{P}_a,\mathcal{P}_b)$ when aligned (c) and when misaligned (d). Blue and red indicate alignment and misalignment respectively.}
\vspace{-0.5cm}
\end{figure}

\begin{figure}
    \centering
    {\includegraphics[trim={0.5cm 7cm 3.7cm 1cm},clip,width=\hsize]{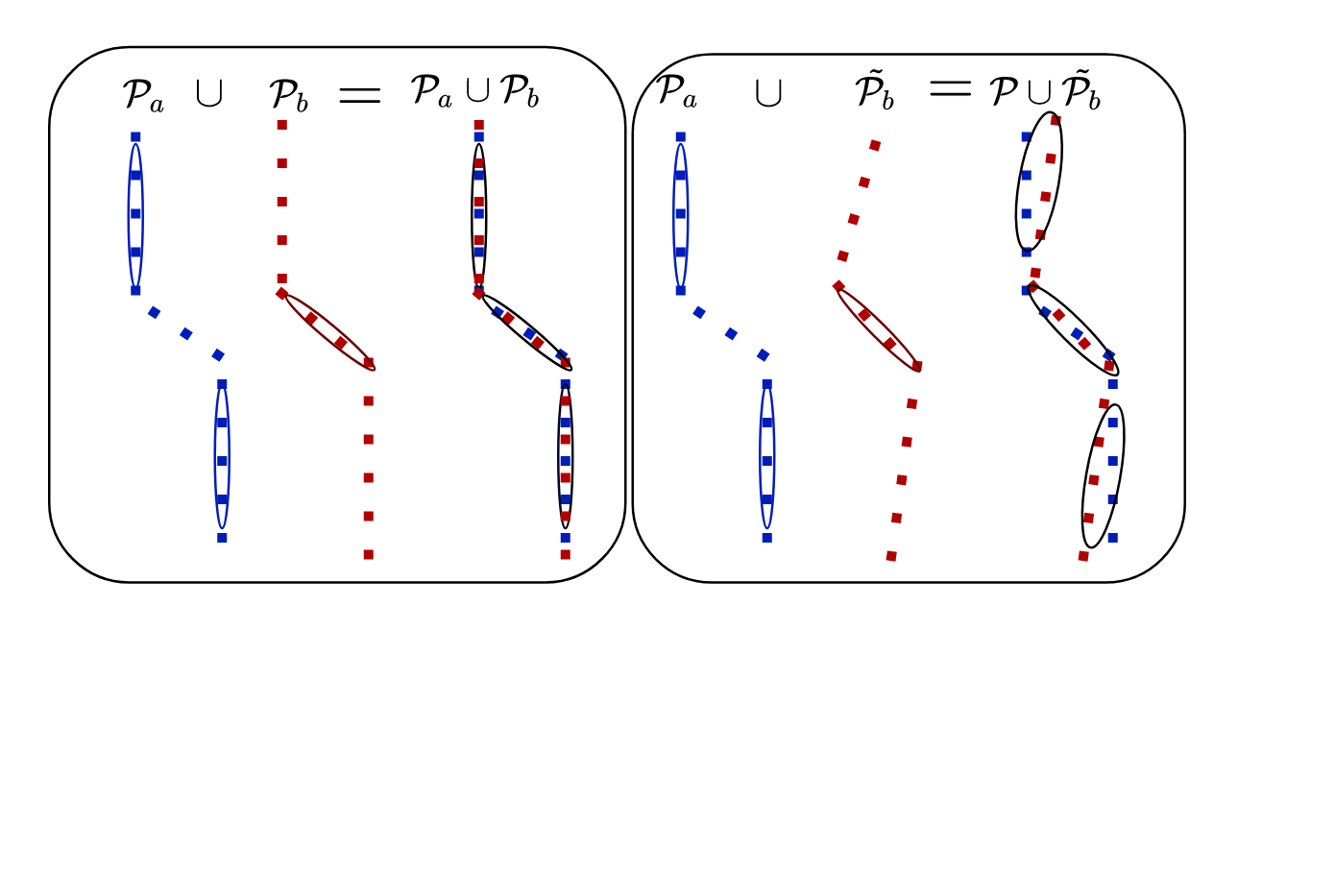}}
    \caption{Example how uncertainty (entropy) is preserved when joining aligned $\mathcal{P}_a\cup\mathcal{P}_b$ (left), but increases when joining misaligned point clouds (right). The entropy for aligned point clouds should be similar to the entropy in the separate point clouds and can be used when quantifying alignment quality.}
    \label{fig:illustration_coral}
    \vspace{-0.5cm}
\end{figure}
Our method is well grounded in information theory and gives an intuitive 
alignment correctness measure. CorAl measures the difference between the average differential entropy in the joint and separate point clouds. For well-aligned point clouds, the joint and the separate point clouds
have similar entropy. In contrast, misaligned point clouds tend to ``blur'' the scene which can be measured as an increase in joint entropy as depicted in \cref{fig:illustration_coral}. By using the separate point clouds to estimate the entropy inherent in the scene, our proposed method can assess quality in a range of different environments.


The contribution of this paper is an intuitive and simple measure of alignment correctness between 
point cloud pairs. We demonstrate how to use this quality measure to train a simple model that detects small alignment errors between point clouds, large errors are not considered in this paper. To train our model, we use previously corrected scans that are assumed to have no alignment error.

We make the following claims: (i) Our proposed method CorAl measures the correctness of point cloud alignment by accounting for the expected scene entropy.
(ii) Our method is accurate in 
a wide range of environments and can generalize well to new environments without retraining.


\section{Related work}
 
Several methods have been used to assess alignment quality in the literature. However, in most cases these methods are used in an ad-hoc manner. Few systematic evaluations of 
their general ability to be used as a classifier to detect aligned vs. misaligned point clouds have been made. 

One well-used alignment measure is the root-mean-squared (RMS) point-to-point distance, truncated by some outlier rejection threshold.  
This is also the function that is minimized by iterative closest point registration~\cite{icp}.
However, this measure has been shown to be highly sensitive to the environment and the choice of the outlier threshold \cite{silva-2005-ga,Almqvist}.
Consequently, this is a poor measure for alignment correctness classification.



One family of methods instead attempts to estimate the alignment uncertainty between point cloud pairs in the form of a covariance matrix~\cite{David_Landry,bengtsson_robot_2003, nieto_scan-slam_2006,prakhya_closed-form_2015}. 
Some use Monte Carlo strategies to estimate  uncertainty by sampling registrations in a region \cite{bengtsson_robot_2003}. This exhaustive search is unpractical in mobile robotics.
Others attempt to estimate uncertainty in closed form using the Hessian \cite{censi-2007-accurate,magnusson-2009-phd,Almqvist}, representing the steepness of the alignment score function around a minimum. These methods assume that the registration has reached a global maximum, which is not necessarily true. Until today, alignment classification based on uncertainty covariance have been less accurate compared to matching score~\cite{Almqvist}.


Almqvist et al.~\cite{Almqvist} explored alignment classifiers based on RMS as well as other existing methods~\cite{rusinkiewicz-2001-fasticp,biber_normal_2003,magnusson-2009-phd,chandran-ramesh_assessing_2007,makadia-2006-fully,silva-2005-ga}, including the NDT score function~\cite{magnusson-2009-phd}, and investigated how to combine the measures with AdaBoost into a stronger classifier.
The classifiers were evaluated on two outdoor data sets, and although their classifiers reached almost 90\,\%  accuracy for the hardest cases on each data set individually, accuracy drops to around 80\,\% when cross-evaluating between the data sets.
In their evaluations, the NDT score function proved to be the best individual measure for alignment assessment. The combined AdaBoost classifier did not have significantly higher accuracy, but reduced parameter sensitivity. 

Liao et al. \cite{fuzzybnb} recently proposed a registration method based on fuzzy clusters, which involves a registration quality assessment.
This fuzzy cluster-based quality assessment (FuzzyQA) compares the similarity of dispersion and disposition of points around fuzzy cluster centers. It has been used to detect if the point clouds are coarsely aligned, Coral instead attempts to detect small alignment errors.

Nobili et al.\cite{8462890} proposed a method to predict alignment risk prior to registration by combining overlap information and an alignment metric. The alignment metric quantifies the geometric constraints in the registration problem. The alignment metric is based on point-to-plane residuals and has been evaluated in structured scenes with planar surfaces, while our method can operate well even in unstructured environments. Additionally, our method seeks to estimate the alignment after registration has been completed to introspectively measure the registration success, as opposed to predicting the risk prior to registration.


Bogoslavskyi et al.~\cite{Bogoslavskyi2017AnalyzingTQ} defined a quality metric based on positive and negative point information, and used it to measure alignment error and cluster three known object types in a controlled experiment. Rather than focusing on objects, our method aims to classify alignment quality of observed scenes in different environments. Additionally, their method operates on range images, which might not be available, while our method operates on unorganized point clouds.

To the best of our knowledge, there is no method for binary point cloud alignment classification
that performs accurately and transfer well to new environments without parameter tuning or retraining.




%




\section{CorAl method}
\label{sec:method}

Our work is inspired by the  Mean-Map-Entropy (MME) measure proposed by Droeschel and Behnke~\cite{8461000} for map quality assessment. MME is based on differential entropy~\cite{article} and measures the randomness of multivariate Gaussian distributions. Droeschel and Behnke  used  MME in absence of accurate ground truth when evaluating map refinement.
As shown in our evaluation, MME cannot be used as a general alignment quality measure as it is also affected by measurement noise, sample density and environment geometry. MME is more affected by changes in the environment compared to CorAl. Hence, the measure is not expected to generalize between, e.g., a structured warehouse and an unstructured outdoor forest environment. We  overcome this effect using dual entropy measurements computed 1) in both point clouds separately and 2) in the joint point cloud. The intuition is that joining two well-aligned point clouds should not introduce additional uncertainty and entropy should remain constant if the point clouds overlap sufficiently.
\subsection{Computing joint and separate entropy}

Our method operates on the dense point clouds $\mathcal{P}_a$, $\mathcal{P}_b$, given in a common fixed world frame, that contain a set of points in the Cartesian space $\bold{p}_k=
\begin{bmatrix}
x & y & z
\end{bmatrix}$.
For later use, we define the joint point cloud $\jointPC=\mathcal{P}_a \cup \mathcal{P}_b$; i.e., all points in $\mathcal{P}_a$ and  $\mathcal{P}_b$ together.

From all points within a radius $r$ around each point $\bold{p}_k$, we compute the sample covariance $\bold{\Sigma(p}_k)$. From the determinant of the sample covariance $\det(\bold{\Sigma}(\bold{p}_k))$ we can then compute the differential entropy as:
\begin{equation}
    \label{eqn:point_entropy}
    h_{i}(\bold{p}_k)=\frac{1}{2}\ln(2\pi e\det(\Sigma(\bold{p}_k)))
\end{equation}
for the point cloud $i=a,b$ that contains $\bold{p}_k$. An example of point clouds colored according to \cref{eqn:point_entropy} can be seen in Fig.~\ref{fig:entropy_src} and \ref{fig:entropy_target}.
The sum of differential entropy for $\mathcal{P}_i$ can then be computed as 
\begin{equation}
    \label{eqn:cloud_etntropy}
    H_i(\mathcal{P}_i)=\sum_{k=1}^{|\mathcal{P}_i|}h_i(\bold{p}_k)
    ,
\end{equation}
where $|\mathcal{P}_i|$ is the number of points in the point cloud $\mathcal{P}_i$. 

Using Eq.~\ref{eqn:cloud_etntropy} we can derive measures of the \emph{separate} and \emph{joint} average differential entropy of two point clouds $\mathcal{P}_a, \mathcal{P}_b$.
\begin{equation}
    \label{eq:Hseparate}
    H_{\mathrm{sep}}=\frac{H_{a}(\mathcal{P}_a)+H_{b}(\mathcal{P}_b)}{|\mathcal{P}_a|+|\mathcal{P}_b|}
    ,
\end{equation}
\begin{equation}
    \label{eq:Hjoint}
    H_{\mathrm{joint}}=\frac{H_j(\mathcal{P}_j)}{|\mathcal{P}_j|}=\frac{H(\mathcal{P}_a\cup\mathcal{P}_b)}{|\mathcal{P}_a|+|\mathcal{P}_b|}
    .
\end{equation}

\begin{figure}
\centering
\subfloat[Top view of first two aligned point clouds $\mathcal{P}_a$(blue) and $\mathcal{P}_b$(red) in the ETH ``stairs'' dataset. 
]{\includegraphics[trim={0cm 13cm 3cm 15cm},clip,width=0.99\hsize,angle=0]{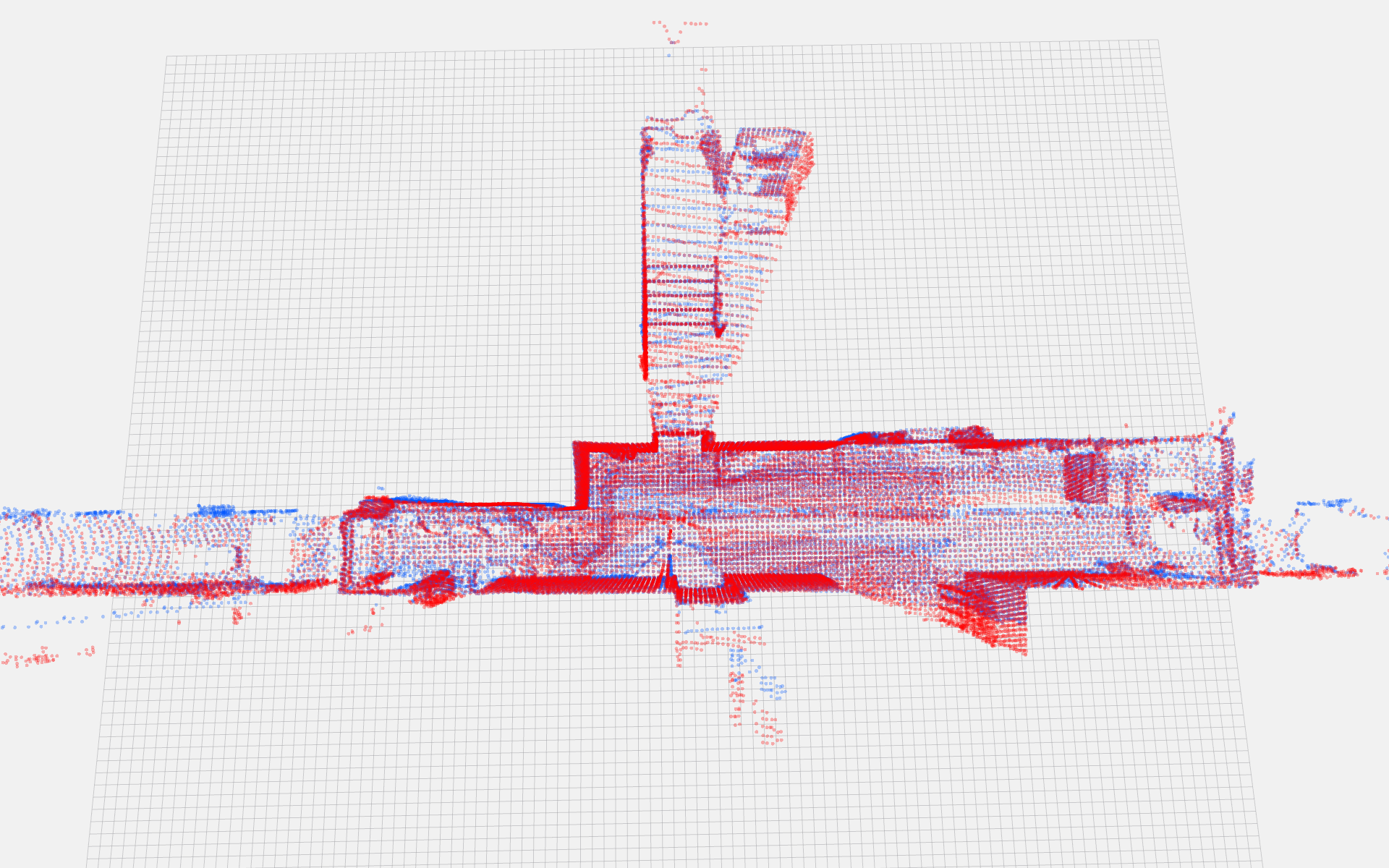}%
\label{fig:top_view_stais}}\\
\vspace{-0.2cm}
\subfloat[][\raggedright CorAl measure $Q(\mathcal{P}_a,\mathcal{P}_b)$ \cref{eqn:EntropyDifference} visualized by color for various (x,y) displacements.]{\includegraphics[trim={0.0cm 0cm 0.0cm 0cm},clip,width=0.49\hsize]{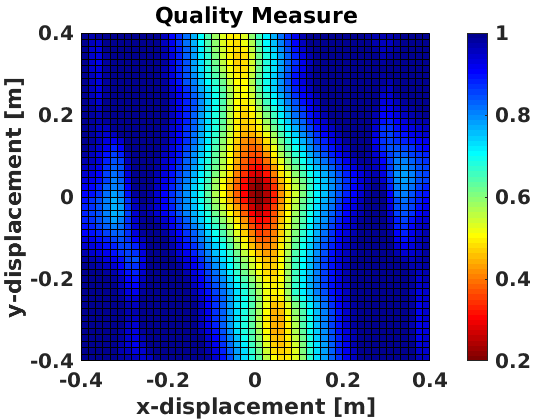}\label{fig:quality_xy_plot}}
\subfloat[][\raggedright CorAl measure $Q(\mathcal{P}_a,\mathcal{P}_b)$ for various (x,$\theta$) displacements.]{\includegraphics[trim={0cm 0cm 0cm 0cm},clip,width=0.49\hsize]{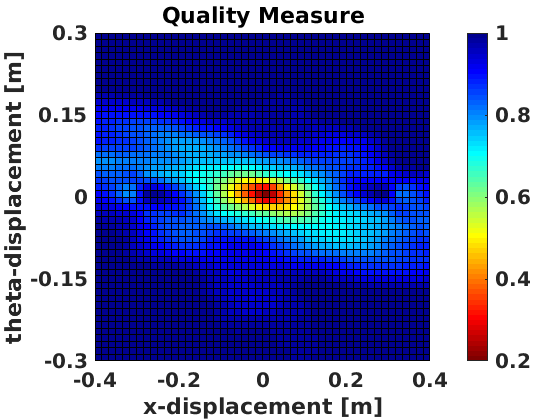}\label{fig:quality_xtheta_plot}}
\caption{
Example of CorAl measure $Q(\mathcal{P}_a,\mathcal{P}_b)$ for various induced (x,y,$\theta$) alignment errors.
$Q$ has a minimum at the true position. The steepness of the surface around the true alignment indicates that CorAl is sensitive to small misalignments.
\label{fig:qualityMeasurePlot}}
\vspace{-0.5cm}
\end{figure}

Our first 
alignment quality measure uses the 
difference between the joint and the separate average differential entropy:

\begin{equation}
    \label{eqn:EntropyDifference}
    Q(\mathcal{P}_a,\mathcal{P}_b)=H_{\mathrm{joint}}(\mathcal{P}_{j})-H_{\mathrm{sep}}(\mathcal{P}_a,\mathcal{P}_b)
    ,
\end{equation}
which can also be given per-point by
\begin{equation}
    \label{eqn:point_quality}
    q_k(\bold{p}_k)=h_{j}(\bold{p}_k)-h_{i}(\bold{p}_k)
    ,
\end{equation}
where the point entropy is evaluated on the joint point cloud $j$ and the separate point cloud $i$ =  ($a$ or $b$) where  $\bold{p}_k$ originates from. An example of point clouds colored by per-point entropy difference according to ~\cref{eqn:point_quality} is depicted in Fig.~\ref{fig:entropy_diff_aligned} and \ref{fig:entropy_diff_misaligned}. Typically, $Q(\mathcal{P}_a,\mathcal{P}_b)$ is close to zero for well-aligned point clouds 
and increases with the alignment error as depicted in \cref{fig:qualityMeasurePlot}, which visualizes the function's surface for position and angular alignment errors around the correct alignment.

Well-aligned point clouds  $\mathcal{P}_a\cup\mathcal{P}_b$ acquired in structured environments have low differential entropy for most query points $\bold{p}_k$. This is reflected by low values for the determinant of the sample covariance. As the determinant can be expressed as the product of the eigenvalues of the sample covariance $\det(\bold{\Sigma(\bold{p}}_k))=\lambda_1\lambda_2\lambda_3$, we see that the measure is sensitive to an increase in the lowest of the eigenvalues when larger eigenvalues are constant. For example, the entropy of points on a planar surface is represented with a flat distribution with two large ($\lambda_1,\lambda_2$) and one small ($\lambda_3$) eigenvalue. Misalignment changes the point distribution in the joint point cloud from flat to ellipsoidal which can be observed as an increase of the smallest eigenvalue $\lambda_3$. This makes the measure sensitive to misalignment of planar surfaces, but generalizes well to other geometries. As shown in the evaluation, the measure can capture discrepancies between point clouds regardless of whether these are due to rigid misalignments or distortions which can can occur when scanning while moving, e.g. because of vibrations or sensor velocity estimation errors. That means that the method can be overly sensitive when used together with a registration method or odometry framework that does not compensate movement distortion or has a low accuracy.


\begin{figure}
\centering
\subfloat[Aligned point clouds: differential entropy distributions are similar.
]{\includegraphics[clip,trim={2.0cm 0cm 2cm 0cm},width=0.49\hsize]{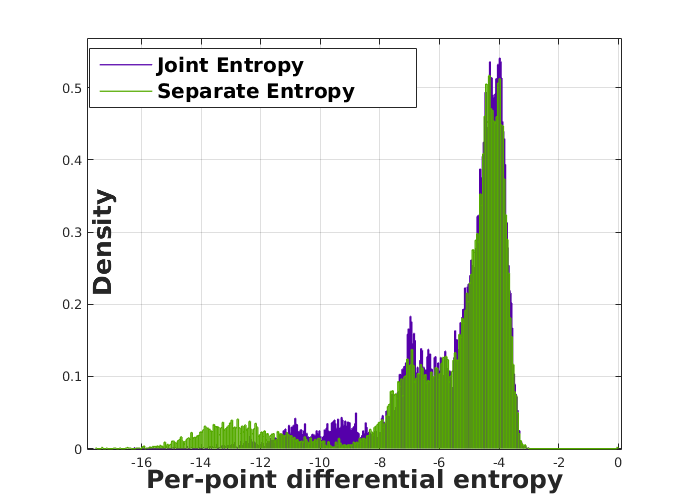}\label{fig:entropy_aligned}}\hfill%
\subfloat[Misaligned point clouds: joint differential entropy is higher.
]{\includegraphics[clip,trim={1.8cm 0cm 1.8cm 0cm},width=0.49\hsize]{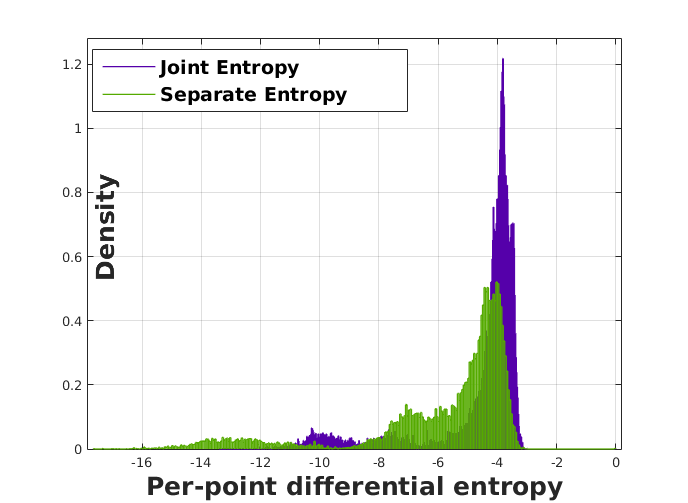}\label{fig:entropy_misaligned}}
\caption{\label{fig:entropypdf}
Probability distribution of per-point entropy \cref{eqn:point_entropy} for joint and separate point clouds when (a) aligned and (b) misaligned. Aligned point clouds have similar joint and separate entropy distributions, while joint entropy is higher compared to separate entropy when point clouds are misaligned.  Joining misaligned point clouds blurs the scene which can be observed by an entropy increase.
An exception can be seen in (a) region $(-12,-8)$ where the entropy increases by joining aligned point clouds.}
\vspace{-0.3cm}
\end{figure}

Overlap is required between point clouds to produce evidence of alignment. For that reason, we classify point clouds with less than 10\% overlap as misaligned. By defining the overlap as all points with a neighbor within $r$ in the other point cloud, non overlapping points have no effect on the quality measure in \cref{eqn:EntropyDifference}.
\subsection{Dynamic radius selection and outlier rejection }
For well aligned point clouds, 
the quality measure is close to zero, meaning that the joint and separate point clouds have similar mean and probability distributions of per-point entropy as depicted in Fig.~\ref{fig:entropy_aligned}.
Unfortunately, the entropy in \cref{eqn:point_entropy} is ill-posed when the determinant
$\det(\bold{\Sigma}(\bold{p}_k))$ is
close to zero and a small increase of the determinant causes a large increase of the entropy. 
Accordingly, the lowest measured entropies can increase (which indicates misalignment) even when joining well aligned point clouds as depicted in~\cref{fig:entropy_aligned}. 
The ill-posed entropies are found where point density is low, typically for solitary points or far from the sensor where the radius $r$ is not large enough to include points that represent the geometry in the environment. The effect of the problem with entropies can be mitigated by maximizing the ratio $Q_s=Q_{misaligned}(\mathcal{P}_a,\mathcal{P}_b)/Q_{aligned}(\mathcal{P}_a,\mathcal{P}_b)$. A larger ratio indicates that the measure is able to discriminate between aligned and misaligned point clouds.
\begin{figure*}[htb!]
  \begin{center}
    \subfloat[]{\includegraphics[trim={0.0cm 2cm 0cm 0cm},clip,width=0.249\hsize]{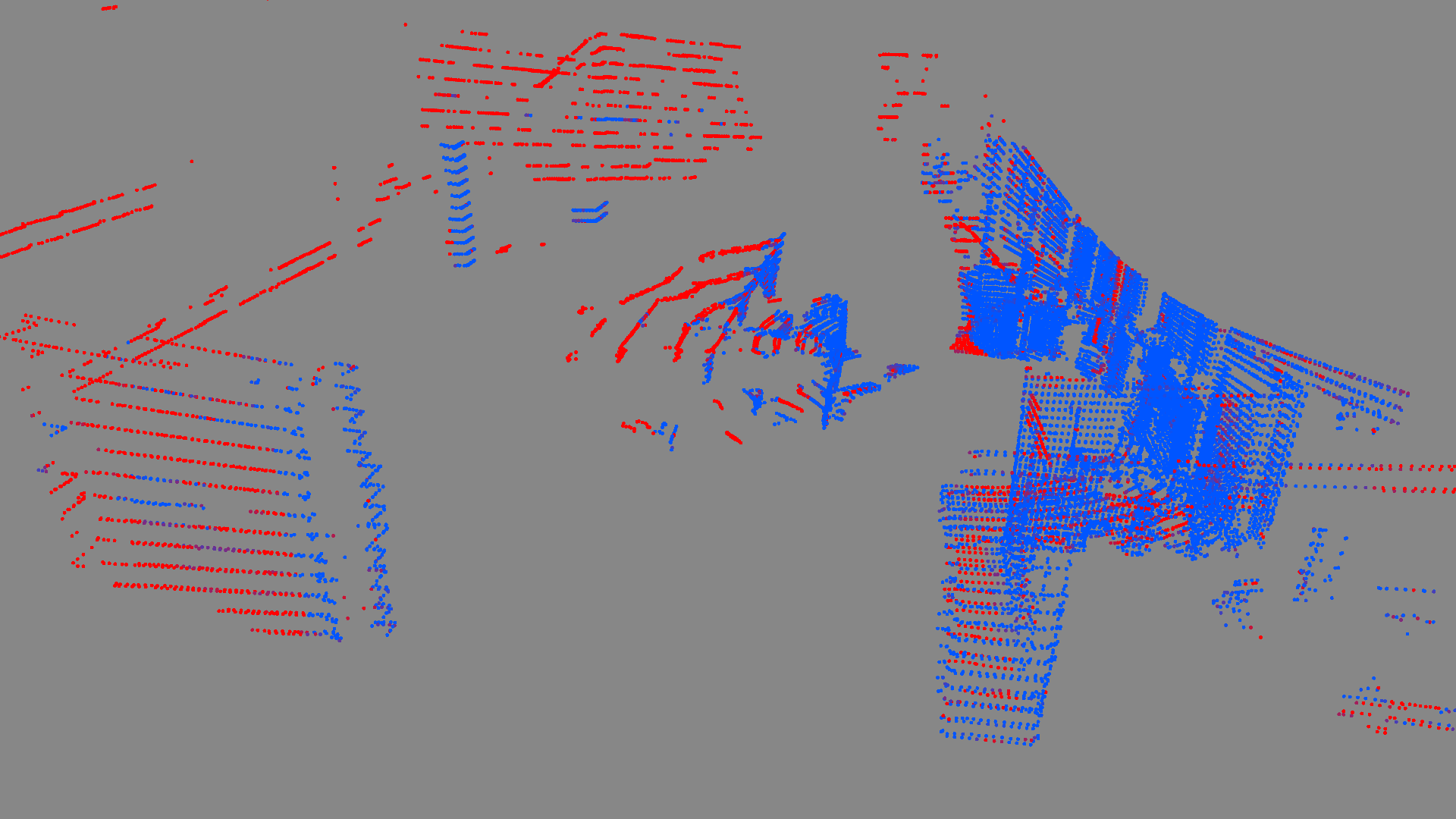}\label{fig:settings_a}}\hfill
    \subfloat[]{\includegraphics[trim={0.0cm 2cm 0cm 0cm},clip,width=0.249\hsize]{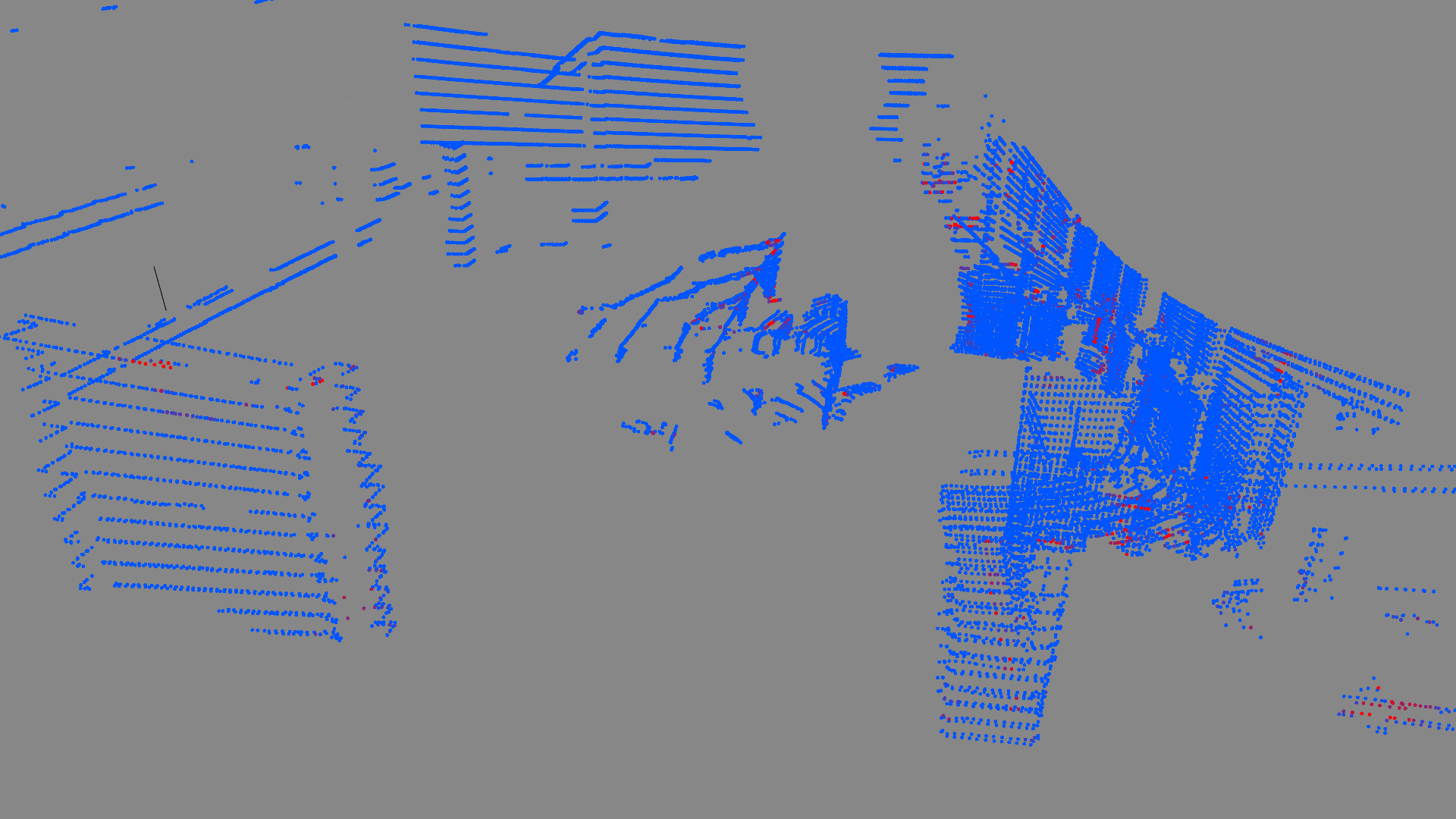}\label{fig:settings_b}}\hfill
    \subfloat[]{\includegraphics[trim={0.0cm 2cm 0cm 0cm},clip,width=0.249\hsize]{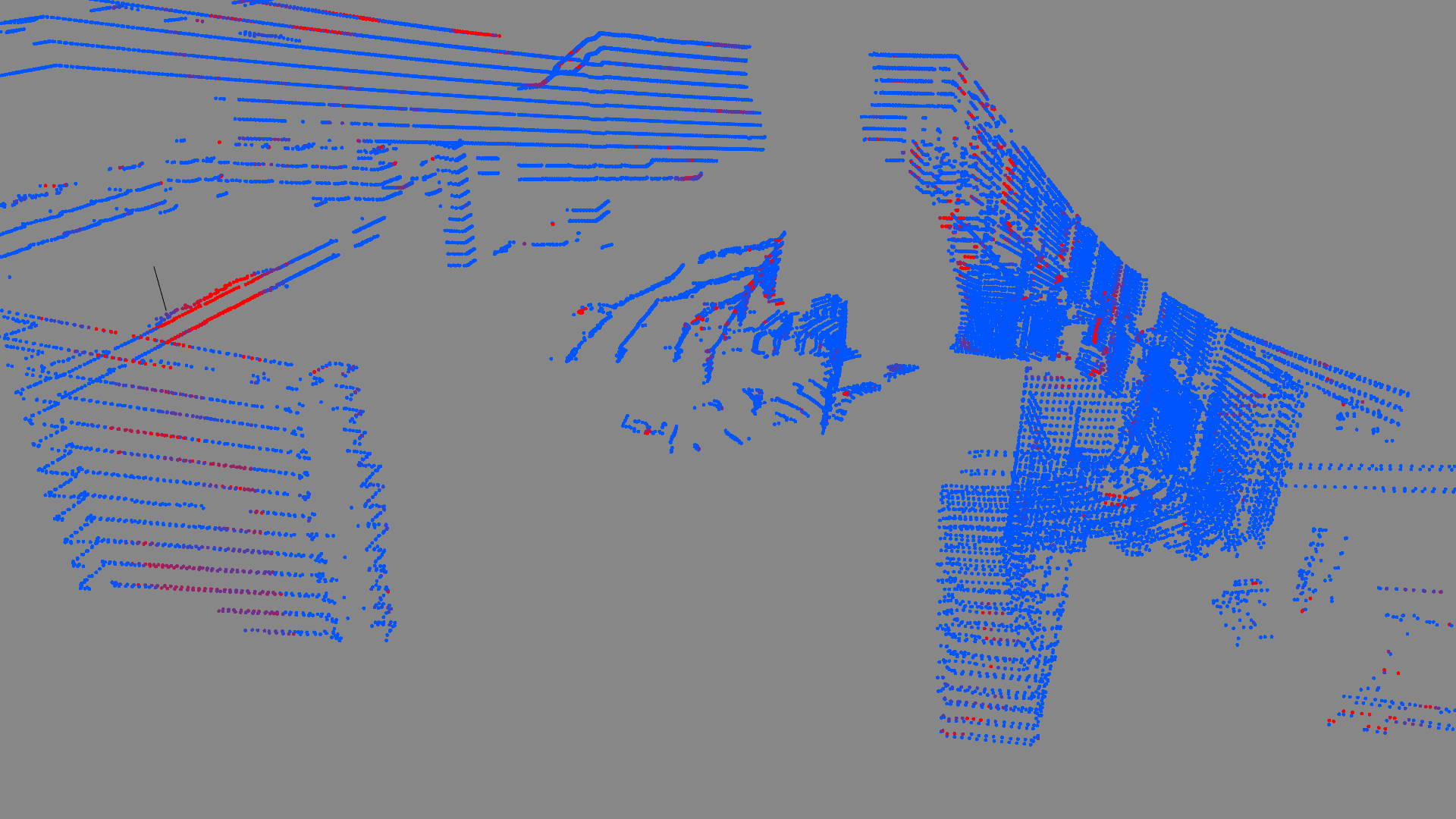}\label{fig:settings_c}}\hfill
    \subfloat[]{\includegraphics[trim={0.0cm 2cm 0cm 0cm},clip,width=0.249\hsize]{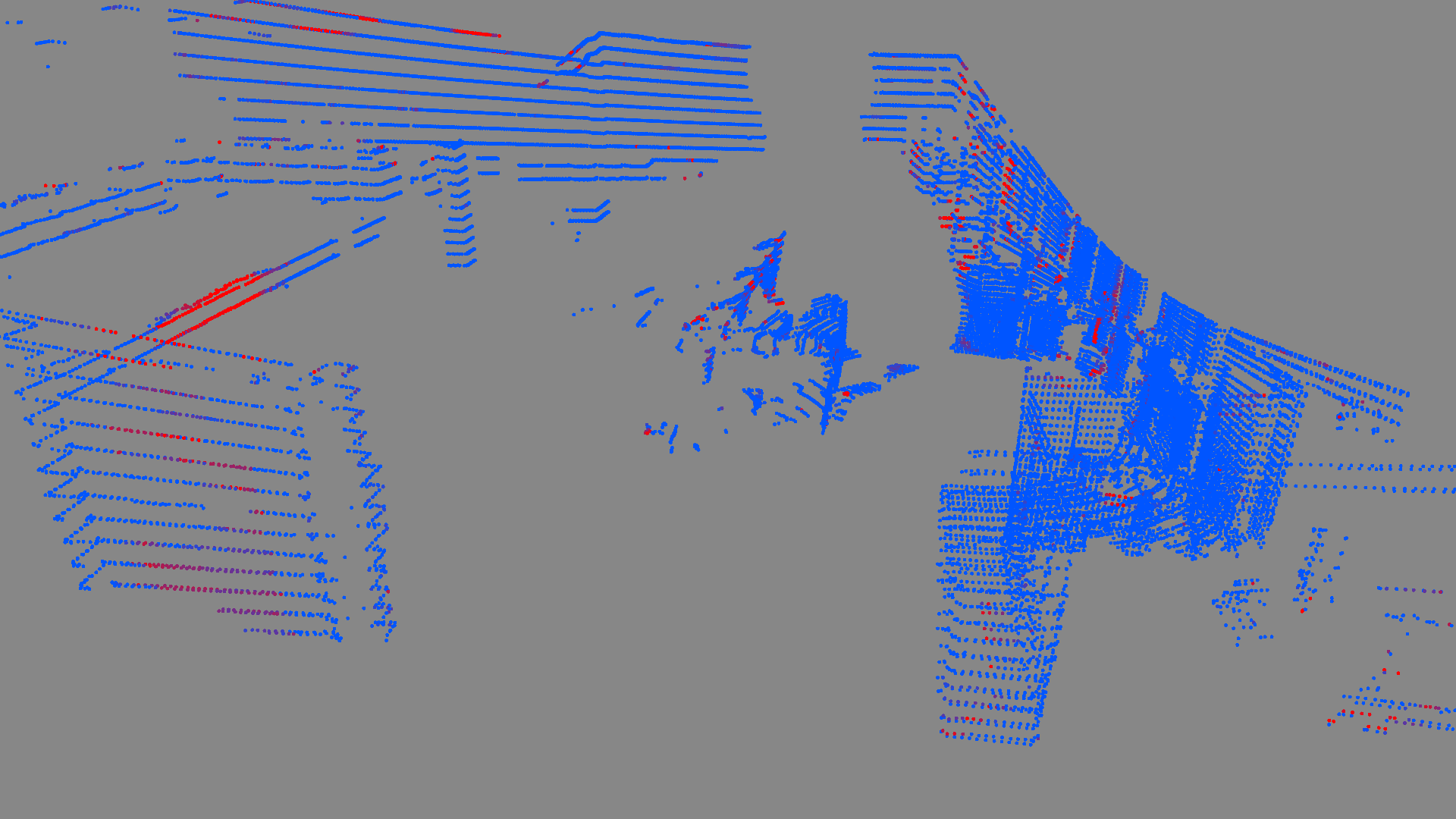}\label{fig:settings_d}}\\
    \vspace{-0.2cm}
     \subfloat[]{\includegraphics[trim={0.0cm 2cm 0cm 0cm},clip,width=0.249\hsize]{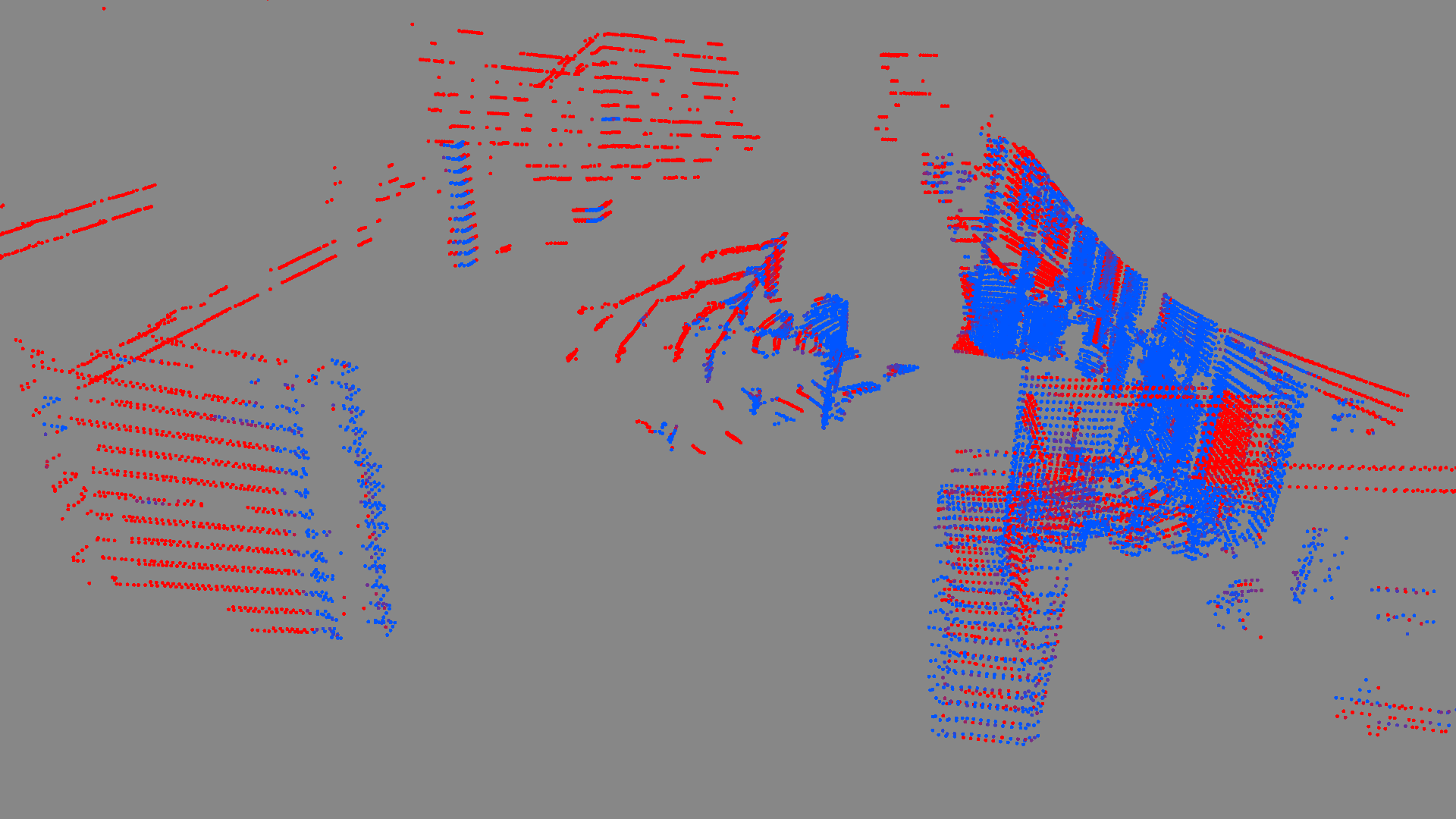}\label{fig:settings_e}}\hfill
    \subfloat[]{\includegraphics[trim={0.0cm 2cm 0cm 0cm},clip,width=0.249\hsize]{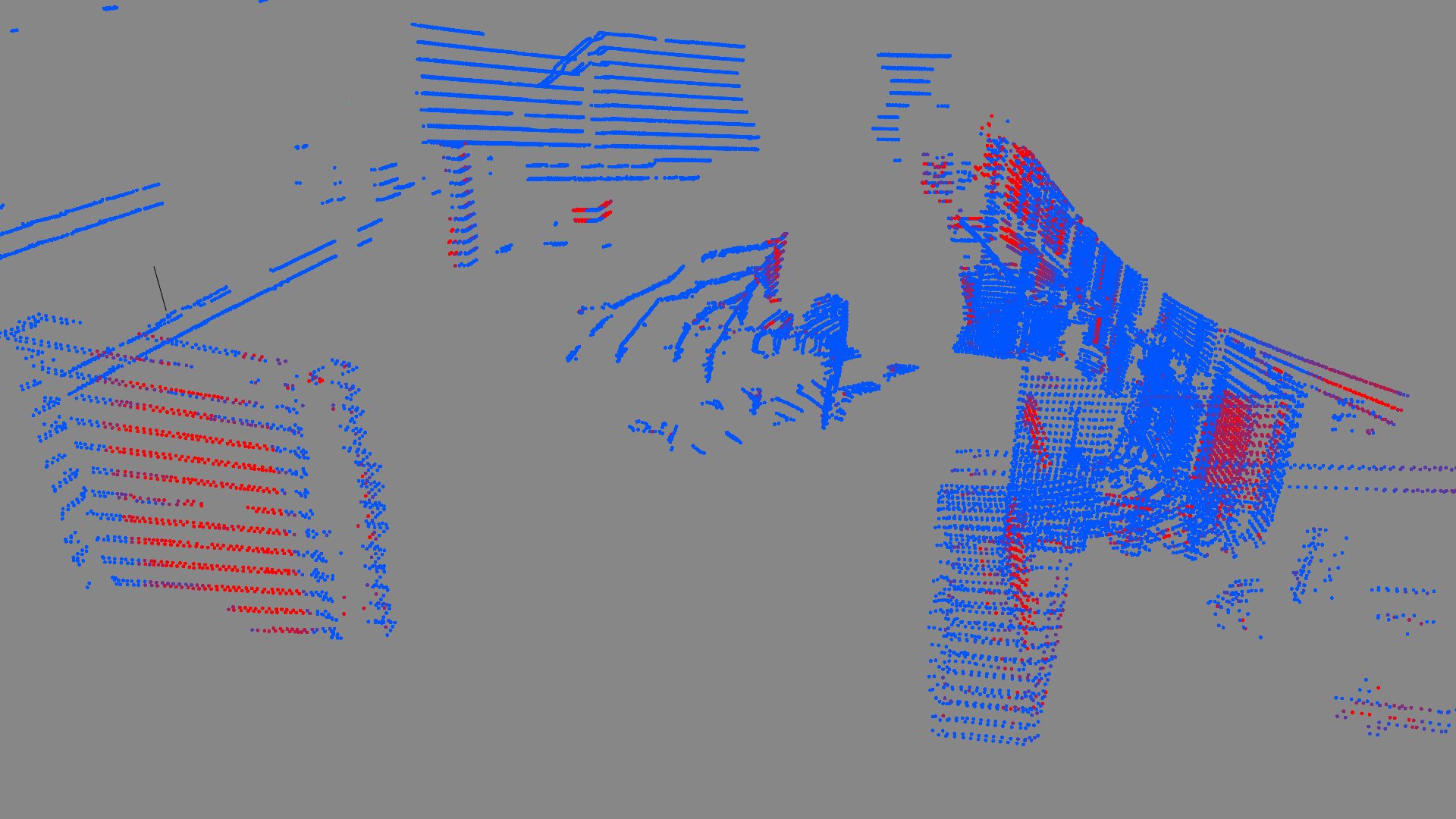}\label{fig:settings_f}}\hfill
    \subfloat[]{\includegraphics[trim={0.0cm 2cm 0cm 0cm},clip,width=0.249\hsize]{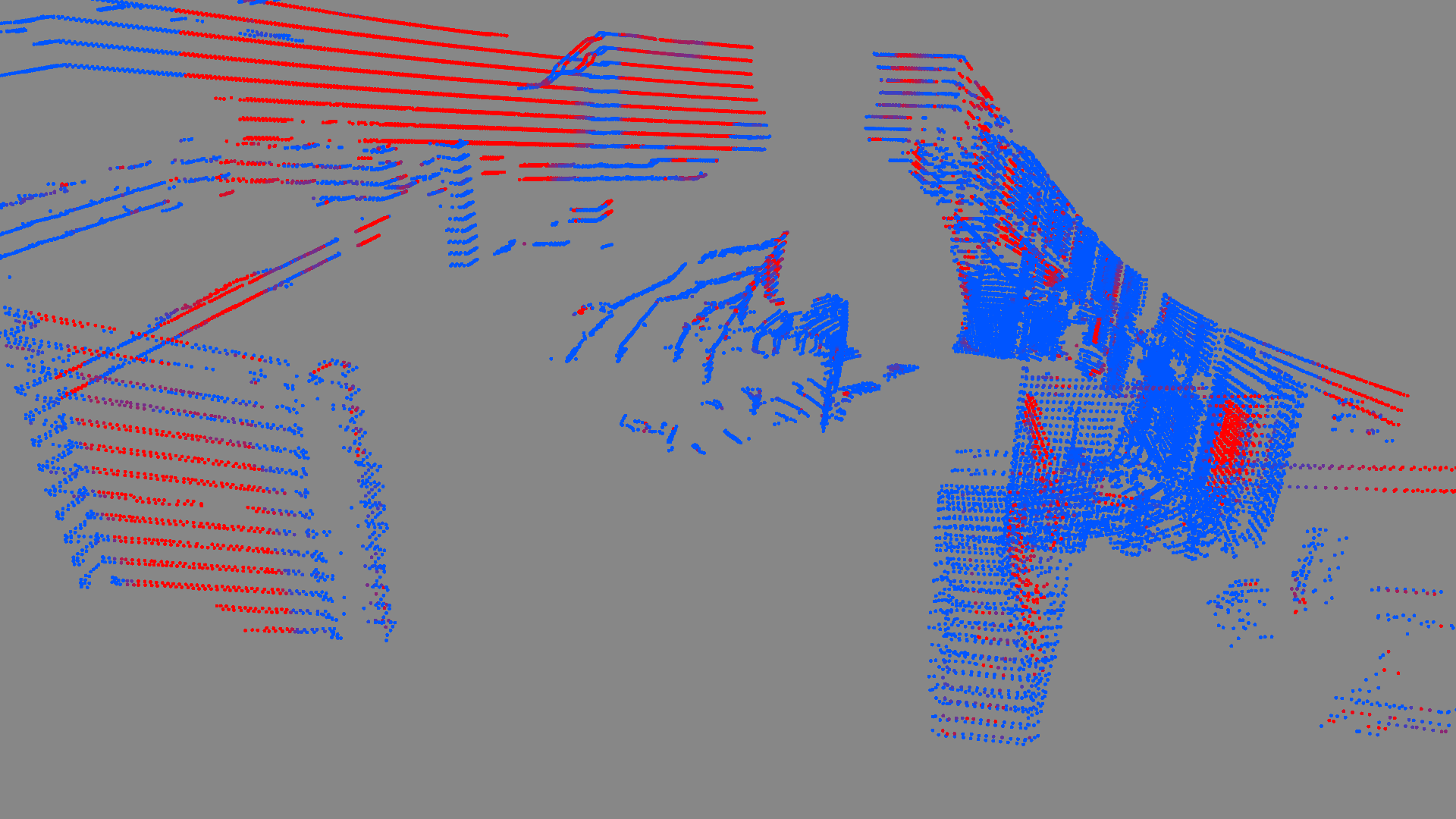}\label{fig:settings_g}}\hfill
    \subfloat[]{\includegraphics[trim={0.0cm 2cm 0cm 0cm},clip,width=0.249\hsize]{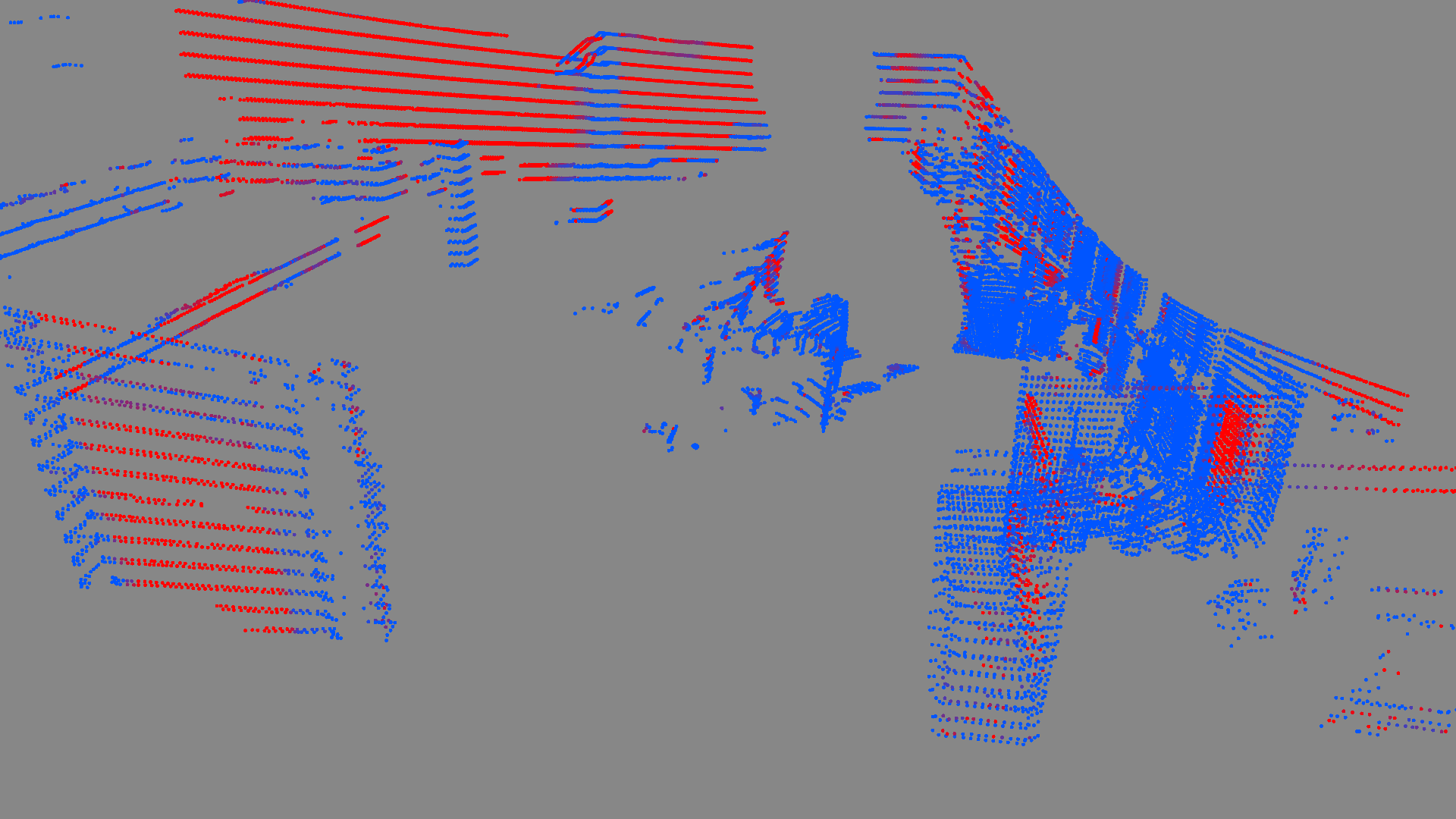}\label{fig:settings_h}}
        	\caption{
        	Joint point cloud colored by per-point quality, $q_k$ ranging from blue (aligned) to red (misaligned). The location of the point cloud origin is highlighted in \cref{fig:orkla}.
        	Columns depict the same parameters for aligned (top) and misaligned (bottom) point clouds.  (a,e): A fixed radius $r=0.3m$ gives $(Q_s=1.46)$. (b,f): radius dynamically adjusted $r_{min}=0.3$m,$\alpha=1.33,r_{max}=0.7$ gives $(Q_s=2.93)$. (c,g):  $\epsilon=10^{-8}$ is added $(Q_s=4.06)$. (d,h): $\Ereject=10\%$ is added ($Q_s=4.3$).  
    	 }\label{fig:modifications}
  \end{center}
  \vspace{-0.3cm}
\end{figure*}
We propose three strategies to address the ill-posed entropies due to variations in sampling density originating from the sensor.

(1): Eq.~\ref{eqn:point_entropy} is modified to   $h_{i}(\bold{p}_k)=\frac{1}{2}\ln(2\pi e\det(\bold{\Sigma(p}_k))+\epsilon)$ where $\epsilon$ limits the lowest possible entropy. 
This make sure that entropy is similar for points distributed along a line and a plane.
The improvement can be seen by comparing  \cref{fig:modifications}(a-b).

(2): Radius $r$ is chosen based on the distance $d$ between the point $\bold{p}_k$ and the sensor location, to account for that point density decrease over distance. The radius is hence selected as: $r=d \sin(\alpha)$ in the range $r_{min} < r < r_{max}$ where $\alpha$ is the vertical resolution of the sensor. 
For other sensor types e.g. RGB-D, the resolution could be chosen similarly according to the angular sensing resolution.
 A dynamic radius enables the quality measure to include more points far from the sensor and correctly detect alignment and misalignment for these as seen in \cref{fig:modifications}(c).

(3): Remove $\Ereject$ percent of points $\bold{p}_k$ with the lowest entropies.
The effect is depicted in \cref{fig:modifications}(d). 


\subsection{Classification}\label{sec:classifier}
We use logistic regression as a model for classification:
\begin{equation}
\begin{split}
    &p=\frac{1}{1+e^{-z}},\\
    &z=\beta_0+\beta_1x_1+\beta_2x_2\\
\end{split}
\end{equation}

\begin{equation}
    \label{eq:classifier}
       y_{pred}= 
\begin{cases}
  \mathrm{aligned} & \mathrm{if} \: p\geq t_h \\
      \mathrm{misaligned} & \mathrm{if} \: p<t_h,
\end{cases}
\end{equation}
where $x_1,x_2$ are input variables (described for each method in \cref{sec:methods}). 
Instead of passing the quality measure $x_1=Q(\mathcal{P}_a,\mathcal{P}_b)$, $H_{sep}$ and $H_{joint}$ are passed separately to $x_1=H_{joint}$ and $ x_2=H_{sep}$.
$\beta_0,\beta_1,\beta_2$ are learned model parameters, $p$ is the class probability and $t_h$ is a class probability threshold and can be adjusted to the application needs. 
For example, in mobile robotics, it is desired that misaligned point clouds are not accidentally reported as aligned (false positives), potentially causing a system failure. In contrast, aligned point clouds classified as misaligned are typically harmless. For that reason, $t_h$ can be increased to reject false positives and hence improve robustness. We used the default threshold $t_h=0.5$.

\section{Evaluation}
\label{sec:benchmark}
 We evaluate an equal portion of aligned and misaligned point clouds. Misaligned point clouds are created by adding an offset for each point cloud pair: an angular offset ($e_{\theta}=0.57^\circ$) around the sensor's vertical axis and a random translational $(x,y)$ offset at a distance ($e_d=0.1$m) from the ground truth. These errors are large enough to be meaningful to detect in various environments, yet challenging to classify.


\subsection{Evaluated methods}\label{sec:methods}
 
The evaluated methods are summarized here together with their most important parameters. 

\paragraph{MME} Mean Map Entropy as proposed by Droeschel and Behnke \cite{8461000} summarized in \cref{eqn:cloud_etntropy}. The parameter is the radius $r$ for associating points.

\paragraph{CorAl (proposed in the paper)} Separate and registered entropy $H_s ,H_j$ as described in~\cref{eq:Hseparate,eq:Hjoint}. Parameters are $r_\mathrm{min}$, $r_\mathrm{max}$ and $\alpha$ to determine nearby points radius, and $\Ereject$ to set outlier rejection ratio and $\epsilon$

\paragraph{CorAl-Median (proposed in the paper)}  $H_s, H_j$ are modified to calculate the median entropy rather than the mean entropy, we hypothesize that this modification can be more robust. The parameters are unchanged.

\paragraph{NDT (point-to-distribution normal-distributions transform)}
The method uses the 3D NDT~\cite{magnusson-2007-jfr} representation similarly to Almqvist~\cite{Almqvist} (NDT3), which constructs a voxel grid over one point cloud, and computes a Gaussian function based on the points in each voxel.
The likelihood of finding the points in $\mathcal{P}_b$, given the NDT representation of $\mathcal{P}_a$, is computed as
\begin{equation}
    s = \frac
    {\sum_{k=1}^{n} \tilde p(\point_k)}
    {n}
    ,
    \label{eqn:ndt_point}
\end{equation}
where $n$ the number of overlapping points, defined as those points (which fall in an occupied NDT voxel, or in a voxel that is a direct neighbor of an occupied voxel) and $\tilde p$ is the probability density function associated with the nearest overlapping NDT-cell.
The most important parameter for NDT is the voxel size $v$ which is set equal to $2*r$ in our evaluation as this makes the sample covariance of NDT cells and entropy computed from points in a similarly large volume. 

\paragraph{Rel-NDT (proposed in the paper)} 
We wanted to investigate if entropy can be used to improve generalization of NDT to different environments. The idea is that environment type is reflected in the average entropy of the scene and can be combined with NDT score to improve classification.
We did this by computing the average entropy
of all NDT-covariances associated with $\point_k$  in the point-likelihood terms  and feed that together with the NDT score \eqref{eqn:ndt_point} to the classifier. 
No additional parameters to NDT are required.

\paragraph{FuzzyQA} FuzzyQA \cite{fuzzybnb} measures the alignment quality by a ratio $\rho=\frac{\mathrm{AFCCD}}{\mathrm{AFPCD}}$, where AFCCD and AFPCD are two indexes describing the points' disposition and dispersion around fuzzy cluster centers. The two point clouds are coarsely aligned if $\rho<1$. However, AFCCD and AFPCD are passed separately to the classifier input $x_1,x_2$.
\paragraph{Input to the classifier}
CorAl, FuzzyQA and Rel-NDT output two decision variables that are passed as input variables  $x_1,x_2$ to the classifier (\ref{sec:classifier}). The other evaluated methods output a single variable $x_1$, and $x_2=0$ is fixed.

\subsection{Qualitative evaluation, live robot data}
\begin{figure}
    \centering
    \includegraphics[width=1\hsize]{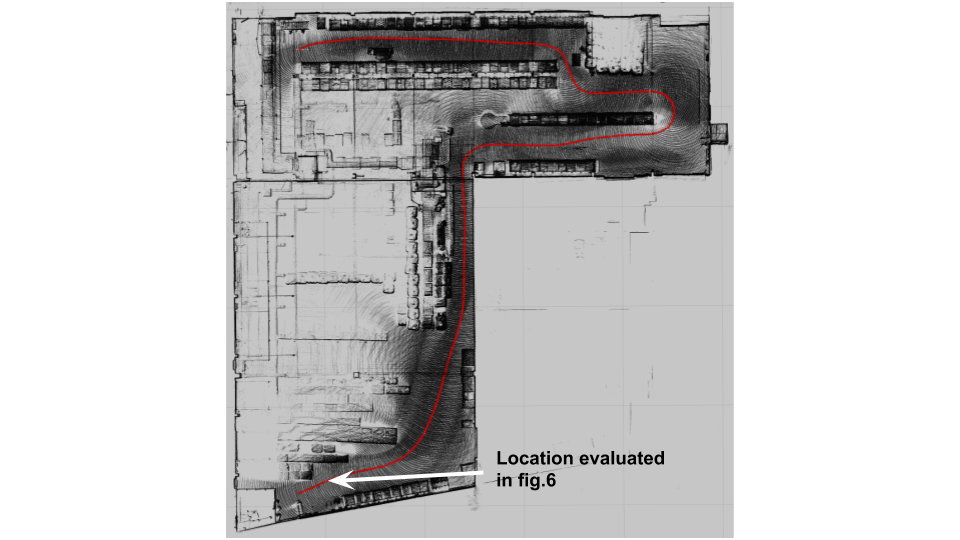}
    \caption{\label{fig:orkla}Data acquired by a truck in a warehouse environment. The sensor trajectory is drawn in red. 
    The environment in the figure is 50 m $\times$ 50 m and the sequence length is 102~m.
    In the first segment of the trajectory, starting at the bottom left, the walls are clearly visible. The final segment is located between aisles where walls are typically out of sight and the sensor observe complex structures such as shelves. 
    The truck traverses over a rough floor with height differences causing vibrations on the sensor.
    }
    \vspace{-0.3cm}
\end{figure}

First, we present qualitative results from real-world data in a structured warehouse environment.
A forklift equipped with a Velodyne HDL-32E spinning laser scanner was manually driven at fast walking speed in the environment depicted in \cref{fig:orkla}. The environment in the sequence varies from large and open with visible walls, to small and narrow between ailes of pallets.
To generate ground-truth alignments for the warehouse dataset, we first aligned the point clouds 
using a scan-to-map approach~\cite{softconstraints}. We then inspected the alignment between subsequent scans and found that at least 40/484 (8.3\%) point clouds were impaired by rigid misalignments or non-rigid distortions from vibrations and motion to the extent that these could be easily visually located.
Alignment classification was then performed on the remaining scans by inducing errors as described in \cref{sec:benchmark}. 
We used the following parameters as they provided a relatively high value of $Q_s$ for the first scan pair in the dataset: $\alpha=0.92^{\circ}$, $\Ereject=0.2$, $r_{min}=0.2$, $r_{max}=1.0$ and voxel size $v=2r_{min}=0.4$.
We found that CorAl-mean, MME and NDT reached an accuracy of $96\%$, $70\%$ and $99\%$ respectively. In this case, NDT performs slightly better than CorAl. 
We believe that CorAl is more sensitive to the typical alignment noise that is still present in the aligned scans. This typical alignment noise introduces a variance in the CorAl score and makes it hard to train a classifier that is sensitive to small misalignment's.
Whether this is desired behavior depends on the application.

\subsection{Quantitative evaluation, ETH benchmark data set}
Our main quantitative evaluation is done
using the public ETH registration dataset \cite{Pomerleau_2012}. This dataset includes 3 sequences in structured (blue) environments (Apartments, ETH Hauptgebaude, Stairs), 3 sequences in semi-structured (brown) environments (Gazebo in summer, Gazebo in winter, Mountain plain) and 2 challenging sequences in unstructured (green) environments (Wood in summer, Wood in autumn). Each sequence contains between 31 and 47 scans acquired from stationary positions.
The dataset contains accurate ground truth positions, required to evaluate the different methods. 
In order to make the evaluation fairer, more realistic and applicable to real applications, we downsample the original, dense, point clouds using a voxel grid of 0.08~m. As the dataset has less variation in sampling density compared to the warehouse dataset, we used a fixed radius $r=0.3$ 
and set $\Ereject=20\%,\epsilon=0$.
NDT voxel size was set equal to the diameter 
$v=2r=0.6$ to create a fair comparison.
\paragraph{Performance}
CorAl has an overall run-time of $0.246\pm 0.095$ seconds per point cloud pair on an Intel Core i7 and depends on the point cloud density. 

\label{sec:eval}

\subsubsection{Separate training}
The first test evaluates the capability to learn classification in a specific type of environment and serves as a reference for further evaluations. The classifiers were trained and evaluated on each sequence separately, using 5-fold cross validation.

\begin{figure}
    \centering
    \includegraphics[trim={2cm 0cm 2cm 0cm},clip,width=\hsize,angle=0]{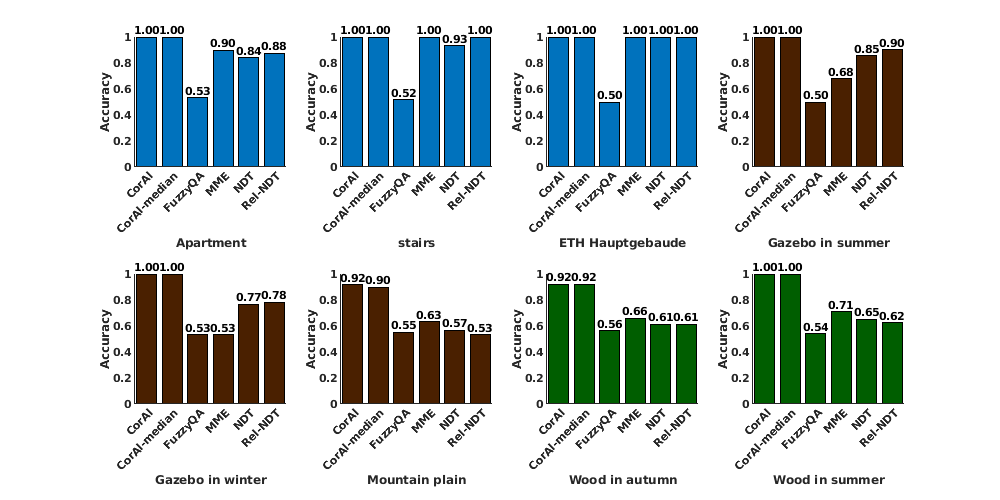}
    \caption{Separate training. 
    The overall accuracy was CorAl: $98\%$, CorAl-median: $98\%$ FuzzyQA: $53\%$ MME: $77\%$ NDT: $78\%$ Rel-NDT: $80\%$}
    \label{fig:ETH_intra_medium}
    \vspace{-0.5cm}
\end{figure}

Results are shown in \cref{fig:ETH_intra_medium}.
We found that all methods except FuzzyQA performed well on the structured environments. We did not expect that FuzzyQA would handle this as it is specifically designed to classify coarse alignment. Surprisingly, even MME scored 90--100\% on the structured environment. This indicates that even naive methods can assess alignment quality in a highly structured 
environment.
In the semi-structured and unstructured sequences, only CorAl and CorAl-median performed well, with consistently $>$90\% accuracy, even in the most challenging sequences. All other methods are only slightly better than random, except for the gazebo sequences. Rel-NDT improves NDT in most cases, however not consistently. We believe this is because entropy alone provides little information about the environment. This is supported by the low overall accuracy of MME.
Both NDT methods performed decently (77--90\%) in the gazebo sequence, indicating that NDT requires at least some structure or surfaces free from foliage to be effective as an alignment correctness measure. 

\subsubsection{Joint training}
\begin{figure}
    \centering
    \includegraphics[trim={2cm 0cm 2cm 0cm},clip,width=\hsize,angle=0]{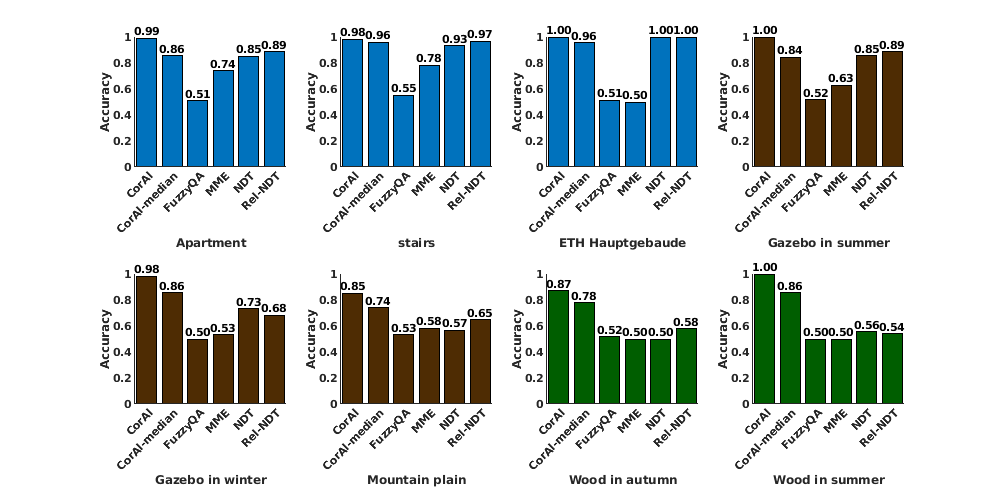}
    \caption{Joint training. Overall accuracy CorAl: $96\%$, Coral-median: $95\%$ FuzzyQA: $52\%$ MME: $60\%$ NDT: $75\%$ Rel-NDT: $78\%$}
    \label{fig:ETH_Joint_medium}
    \vspace{-0.5cm}
\end{figure}

The second test evaluates how the methods are able to learn alignment classification when trained in a variety of environments. To do that, the methods need to be versatile. Training was performed on all the ETH sequences, evaluation was then performed on each sequence individually. The results are shown  in \cref{fig:ETH_Joint_medium}.
The accuracy of all classifiers decreased compared to the previous test. CorAl performed best, with accuracy  85--100\% in all cases. CorAl-median reached a slightly lower accuracy compared to CorAl. Rel-NDT performed better than NDT in most cases, however not consistently. The generally high accuracy of CorAl indicates that it is possible to find general parameters that makes the method valid in various environments.

\subsubsection{Generalization to unseen environments}
\begin{figure}
    \centering
    \includegraphics[trim={2cm 0cm 2cm 0cm},clip,width=\hsize,angle=0]{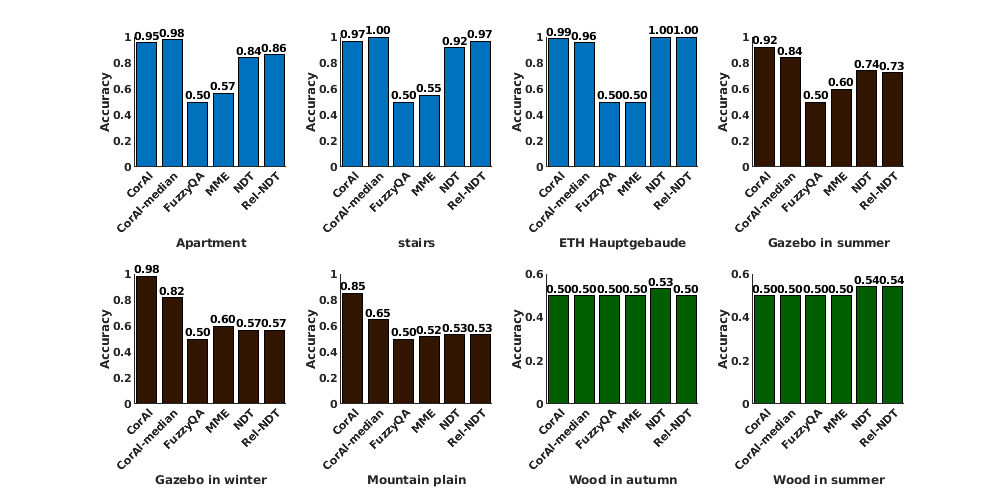}
    \caption{Evaluation on unseen environments. Overall accuracy: $83\%$, CorAl-median: $79\%$ FuzzyQA: $50\%$ MME: $54\%$ NDT: $72\%$ Rel-NDT: $72\%$, In structured and semi-structured environments: $95\%$, Coral-median: $88\%$ FuzzyQA: $50\%$ MME: $56\%$ NDT: $78\%$ Rel-NDT: $79\%$.}
    \label{fig:ETH_generalization_medium}
    \vspace{-0.5cm}
\end{figure}

The final test evaluates how classifiers perform in environments with different characteristics than those observed in the training set. We trained and evaluated on different sequences and environments. The 3 structured environments were used for training and the remaining 5 (semi-structured and unstructured) were used for evaluation and vice versa. The classification accuracy is depicted in \cref{fig:ETH_generalization_medium}.
When trained on structured  and evaluated on semi-structured environments, CorAl performed accurately(85--98\%) and other methods performed close to random except NDT for Gazebo summer ($74\%$) No method generalized well from structured to unstructured environments. 
On the other hand, learning from semi-structured and unstructured environments was enough to afford very high accuracy in structured environments with CorAl -- very close to what was attained with joint training on all sequences. 
The previous joint evaluation show that it's possible to train a model that is simultaneously accurate in all environment types. For that reason, we believe that the reason the classifier trained in a structured environment does not generalize to an unstructured environment is that the model overfits when not using sufficiently diverse and challenging data.

\section{Conclusions}
In this paper we introduced CorAl, a principled and intuitive measure of alignment correctness between point clouds. 
Using dual entropy measurement that compares the expected entropy found in the separate point clouds with the actual entropy, 
CorAl can measure point cloud alignment correctness and substantially outperforms previous methods when evaluated on a public data set. 
Specifically, we were able to use CorAl to train a classifier based on logistic regression that is simultaneously accurate in a diverse range of environments. 
Our experiments show that our method generalizes well from (i) unstructured and semi-structured to structured environments, and (ii) from structured to semi-structured.
None of the evaluated methods generalized well from structured to unstructured environments.
Therefore, we conclude it is possible to train a general and accurate alignment classifier given that training data is sufficiently diverse.
Relatively modest results $96\%$ was achieved on live data. We think that the poor quality of the ground truth (obtained by lidar odometry and manual inspection) causes high variance in the CorAl score. The score is sensitive to small misalignment's, therefore a higher quality ground truth is required to make a fair evaluation.
We believe that CorAl per-point quality and classification can be a useful tool for alignment evaluation and can improve robustness in various perception tasks by serving as a fault detection step.

In the future we will investigate how to automatically learn sensor specific parameters or use the range image to find neighbouring points for covariance computation. This could address variations in point density owed to different sensors and environment scales.


\addtolength{\textheight}{-5cm}   
\bibliographystyle{IEEEtran}
\bibliography{references}
\end{document}